\documentclass{bmcart}

\usepackage{amsthm,amsmath}
\usepackage{amsfonts}
\usepackage{physics}
\usepackage{bm}
\usepackage[utf8]{inputenc} 
\usepackage{lipsum}
\usepackage[ruled]{algorithm2e}
\SetKwRepeat{Do}{do}{while}%
\usepackage{booktabs}       
\usepackage{multirow}
\usepackage{siunitx}
\usepackage{hyperref}
\usepackage{graphicx}
\usepackage{caption2}
\usepackage{subfigure}
\usepackage{threeparttable}
\usepackage{float}
\graphicspath{ {./} }
\usepackage{xcolor}

\newcommand{\mystar}[1]{${#1}^{\star}$}

\startlocaldefs
\endlocaldefs

\begin{document}

\begin{frontmatter}

\begin{fmbox}
\dochead{Research}


\title{Gumbel-softmax-based Optimization: A Simple General Framework for Optimization Problems on Graphs}



\author[
   addressref={aff1},
   email={leofrancescozn@gmail.com}
]{\inits{YXL}\fnm{Yaoxin} \snm{Li}}
\author[
   addressref={aff1},
   email={jing.liu@mail.bnu.edu.cn}
]{\inits{JL}\fnm{Jing} \snm{Liu}}
\author[
   addressref={aff1},
   email={gzhlin@mail.bnu.edu.cn}
]{\inits{GZL}\fnm{Guozheng} \snm{Lin}}
\author[
   addressref={aff2},
   email={houyueyuan@caiyunapp.com}
]{\inits{YYH}\fnm{Yueyuan} \snm{Hou}}
\author[
   addressref={aff1},
   email={moumuyun@gmail.com}
]{\inits{MYM}\fnm{Muyun} \snm{Mou}}
\author[
   addressref={aff1},
   corref={aff1},
   email={zhangjiang@bnu.edu.cn}
]{\inits{JZ}\fnm{Jiang} \snm{Zhang}}


\address[id=aff1]{
  \orgname{School of Systems Science, Beijing Normal University}, 
  \street{No.19, Xinjiekouwai St, Haidian District},                     %
  \postcode{100875}                                
  \city{Beijing},                              
  \cny{P.R.China}                                    
}

\address[id=aff2]{%
  \orgname{ColorfulClouds Tech},
  \street{No.04, Building C, 768 Creative Industrial Park, Compound 5A, Xueyuan Road, Haidian District},
   \postcode{100083}
  \city{Beijing},
  \cny{P.R.China}
}


\begin{artnotes}
\note[id=n1]{Equal contributor} 
\end{artnotes}

\end{fmbox}


\begin{abstractbox}

\begin{abstract} 

In computer science, there exist a large number of optimization problems defined on graphs, that is to find a best node state configuration or a network structure such that the designed objective function is optimized under some constraints. However, these problems are notorious for their hardness to solve because most of them are NP-hard or NP-complete. Although traditional general methods such as simulated annealing (SA), genetic algorithms (GA) and so forth have been devised to these hard problems, their accuracy and time consumption are not satisfying in practice. In this work, we proposed a simple, fast, and general algorithm framework based on advanced automatic differentiation technique empowered by deep learning frameworks. By introducing Gumbel-softmax technique, we can optimize the objective function directly by gradient descent algorithm regardless of the discrete nature of variables. We also introduce evolution strategy to parallel version of our algorithm. We test our algorithm on three representative optimization problems on graph including modularity optimization from network science, Sherrington-Kirkpatrick (SK) model from statistical physics, maximum independent set (MIS) and minimum vertex cover (MVC) problem from combinatorial optimization on graph. High-quality solutions can be obtained with much less time consuming compared to traditional approaches.
\end{abstract}




\end{abstractbox}
%

\end{frontmatter}



\section*{Introduction}
\label{section: intro}
In computer science, there exist a large number of optimization problems defined on graphs, e.g., maximal independent set (MIS) and minimum vertex cover (MVC) problems \cite{karp1972reducibility}. In these problems, one is asked to give a largest (or smallest) subset of the graph under some constraints. In statistical physics, finding the ground state configuration of spin glasses model where the energy is minimized is another type of optimization problems on specific graphs \cite{mezard1987spin}. Obviously, in the field of network science there are a great number of optimization problems defined on graphs abstracted from real-world networks. For example, modularity maximization problem \cite{newman2006modularity} asks to specify which community one node belongs to so that the modularity value is maximized. In general, the space of possible solutions of mentioned problems is typically very large and grows exponentially with system size, thus impossible to solve by exhaustion. 

There are many algorithms for optimization problem. Coordinate descent algorithms (CD) which based on line search is a classic algorithm and solve optimization problems by performing approximate minimization along coordinate directions or coordinate hyperplanes \cite{Wright2015}. However, it does not take gradient information into optimizing process and can be unstable on unsmooth functions. Particle swarm optimization (PSO) is another biologically derived algorithm that can be effective for optimizing a wide range of functions \cite{kennedy95particle}.  It is highly dependent on stochastic processes, and it does not take advantage of gradient information either. Other widely-used methods such as simulated annealing (SA) \cite{kirkpatrick1983optimization}, genetic algorithm (GA) \cite{davis1991handbook}, extremal optimization (EO) \cite{boettcher2000nature} are capable of solving various kinds of problems. However, when it comes to combinatorial optimization problems on graphs, these methods usually suffer from slow convergence and are limited to system size up to thousand. Although there exist many other heuristic solvers such as local search \cite{andrade2012fast}, they are usually domain-specific and require special domain knowledge.

Fortunately, there are other optimization methods based on gradient descent that are able to work without suffering from these drawbacks. However, these gradient-based methods require the gradient calculation has to be designed manually throughout the optimization process for each specific problems, thereafter, they lack flexibility and generalizability.

Nowadays, with automatic differentiation technique \cite{paszke2017automatic} developed in deep learning area, gradient descent based methods have been renewed. Based on computational graph and tensor operation, this technique automatically calculates the derivative so that back propagation can work more easily. Once the forward computational process is well defined, the automatic differentiation framework can automatically compute the gradients of all variables with respect to the objective function. 

Nevertheless, there exist combinatorial optimization problems on graphs whose objective functions are non-differentiable, therefore cannot be solved by using automatic differentiation technique. Some other techniques developed in reinforcement learning area seek to solve the problems directly without training and testing stages. For example, REINFORCE algorithm \cite{williams1992simple} is a typical gradient estimator for discrete optimization. Recently, reparameterization trick, which is a competitive candidate of REINFORCE algorithm for estimating gradient, is developed in machine learning community. For example, Gumbel-softmax \cite{gumbel,concrete} provides another approach for differentiable sampling. It allows us to pass gradients through sampling process directly. It has been applied on various machine learning problems\cite{gumbel,concrete}. 

With reparameterization trick such as Gumbel-softmax, it is possible to treat many discrete optimization problems on graphs as continuous optimization problems \cite{Andreasson2007} and apply a series of gradient descent based algorithms \cite{Avraamidou2020}. Although these reinforcement learning and reparameterization tricks provide us a new way to solve discrete problems, when it comes to complicated combinatorial optimization problems on large graphs, the performances of these methods are not satisfying because they often stuck with local optimum.

Nowadays, a great number of hybrid algorithms taking advantage of both gradient descent and evolution strategy have shown their effectiveness over optimization problems \cite{Zidani2020,Rocha2020} such as function optimization. Other population based algorithms \cite{Yildiz2012} also show potential to work together with gradient based methods to achieve better performance.

In this work, we present a novel general optimization framework based on automatic differentiation technique and Gumbel-softmax, including Gumbel-softmax optimization (GSO) \cite{liu2019gumbel} and Evolutionary Gumbel-softmax optimization (EvoGSO). The original Gumbel-softmax optimization algorithm applies Gumbel-softmax reparameterization trick on combinatorial problems on graphs directly to convert the original discrete problem into a continuous optimization problem such that the gradient decent method can be used. The batched version of GSO algorithm improves the results by searching the best solution in a group of optimization variables undergoing gradient decent optimization process in a parallel manner. The evolutionary Gumbel-softmax optimization method builds a mixed algorithm that combines the batched version of GSO algorithm and evolutionary computation methods. The key idea is to treat the batched optimization variables - the parameters as a population such that the evolutionary operators, e.g. substitution, mutation, and crossover can be applied. The introduction of evolutionary operators can significantly accelerate the optimization process.

We first introduce our method proposed in \cite{liu2019gumbel} and then the improved algorithm: Evolutionary Gumbel-softmax (EvoGSO). Then we give a brief description of three different optimization problems on graphs and specify our experiment configuration, followed by main results on these problems, compared with different benchmark algorithms. The results show that our framework can achieve competitive optimal solutions and also benefit from time consumption. Finally we give some concluding remarks and prospect of future work.

\section*{The proposed algorithm}
\label{sec:The proposed algorithm}
In \cite{liu2019gumbel} we proposed Gumbel-softmax optimization (GSO), a novel general method for solving combinatorial optimization problems on graphs. Here we briefly introduce the basic idea of GSO and then introduce our improvement: Evolutionary Gumbel-softmax optimization (EvoGSO).

\subsection*{Gumbel-softmax optimization (GSO)}
Considering an optimization problems on graph with $N$ nodes, each node can take $K$ different values, i.e., selected or non-selected for $K=2$. Our goal is to find configuration $\mathbf{s}=(s_1, s_2, \cdots, s_N)$ that minimizes the objective function. 
Suppose we can sample from all allowed solution space easily, we want those configurations with lower objective function to have higher probabilities $p(\mathbf{s})$. Here, $p(\mathbf{s})$ is the joint distribution of solutions, which is the key for the optimization.
There are a large number of methods to specify the joint distribution, among which the mean field factorization is the simplest one. That is, we factorize the joint distribution of solutions into the product of $N$ independent categorical distributions \cite{wainwright2008graphical}, which is also called naive mean-field in physics:
\begin{equation}
  p(s_1, s_2, \cdots, s_N) = \prod_{i=1}^N p_{\theta}(s_i).
  \label{eq:jointdistribution}
\end{equation}
and the marginal probability $p(s_i)\in [0,1]^K$ can be parameterized by a set of parameters $\theta_i$ which is easily generated by Sigmoid or softmax function. 

It is easy to sample a possible solution $\mathbf{s}$ according to Equation~\ref{eq:jointdistribution} and then evaluate the objective function $E(\mathbf s;\pmb \theta)$. However, due to the non-differentiable nature of sampling, we cannot estimate the gradients of $\pmb{\theta}$ unless we resort to Monte Carlo gradient estimation techniques such as REINFORCE \cite{williams1992simple}. Gumbel-softmax \cite{gumbel}, also known as concrete distribution \cite{concrete} provides an alternative approach to tackle the difficulty of non-differentiability. Consider a categorical variable $s_i$ that can take discrete values $s_i \in \{1,2,\cdots, K\}$. This variable $s_i$ can be parameterized as a \textit{K}-dimensional vector $(p_1, p_2, \cdots, p_K)$ where $\theta_i$ is the probability that $\theta_i=p(s_i=r), r=1, 2, \cdots, K$. Instead of sampling a hard one-hot vector, Gumbel-softmax technique gives a \textit{K}-dimensional sampled vector where the \textit{i}-th entry is
\begin{equation}
\hat{p}_{i}=\frac{\exp \left(\left(\log \left(p_{i}\right)+g_{i}\right) / \tau\right)}{\sum_{j=1}^{K} \exp \left(\left(\log \left(p_{j}\right)+g_{j}\right) / \tau\right)} \quad \text { for } i=1,2, \cdots, K,
\label{eq:gumbel}
\end{equation}
where $g_i \sim \text{Gumbel}(0,1)$ is a random variable following standard Gumbel distribution and $\tau$ is the temperature parameter. Notice that as $\tau \rightarrow 0$, the softmax function will approximate $\text{argmax}$ function and the sampled vector will approach a one-hot vector. And the one-hot vector can be regarded as a sampled solution according to the distribution $(p_1,p_2,\cdots,p_K)$ because the unitary element will appear on the $i^{th}$ element in the one-hot vector with probability $p_i$, therefore, the computation of Gumbel-softmax function can simulate the sampling process. Furthermore, this technique allows us to pass gradients directly through the ``sampling'' process because all the operations in Equation \ref{eq:gumbel} are differentiable. In practice, it is common to adopt a annealing schedule from a high temperature $\tau$ to a small temperature. 

\begin{figure}[!htbp]
	\centering
	\includegraphics[width=0.98\linewidth]{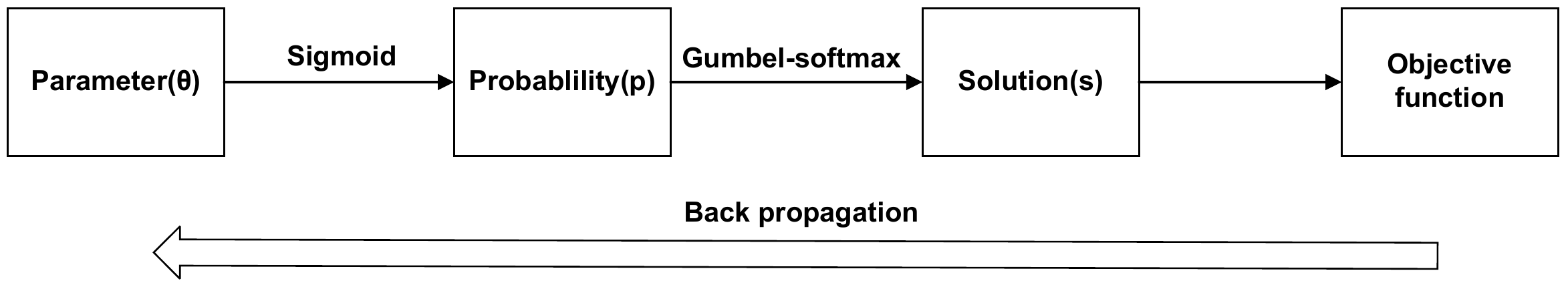}
	\caption{Process of Gumbel-softmax optimization}
	\label{fig:visio}
\end{figure}

In a concise manner, we randomly initialize a series of learnable parameters $\pmb{\theta}$ which are the variables for optimization and the probabilities $\pmb{p}$ are generated by Sigmoid function over these parameters. Then we sample from $\pmb{p}$ with Gumbel-softmax technique to get solutions and calculate objective function. Finally, we run back propagation algorithm to update parameters $\pmb{\theta}$. The whole process is briefly demonstrated in Figure \ref{fig:visio}.

\subsection*{Parallel version of GSO}
We point out that our method can be implemented in parallel on GPU: $N_{\text{bs}}$ different learnable parameters $\pmb{\theta}$ can form a group which is called a batch. These parameters are initialized and optimized simultaneously. So we have $N_{\text{bs}}$ candidate solutions in a batch instead of one. When the optimizing procedure is finished, we select the solution with the best performance from this batch. In such a way, we can take full advantage of GPU acceleration and obtain better results more likely. 

The whole process of optimization solution is presented in Algorithm~(\ref{GSO}).

\begin{algorithm}[!htbp]
  \KwIn{Problem size $N$, batch size $N_{\text{bs}}$, learning rate $\eta$, and Graph $\mathcal{G}$ for optimization.}
	\KwOut{solution with the best performance}
  Initialize $\bm \theta = (\theta_1, \theta_2, \cdots, \theta_N) \in \mathbb{R}^{N_{bs}\times N \times K}$\;
	\Repeat{Convergence}
	{
		$\mathbf{s} \leftarrow \text{Gumbel-softmax sampling from } p_{\bm \theta} (p_{\bm \theta}=\text{Sigmoid}(\bm \theta))$\;
		$E \leftarrow E(\mathbf s;\bm \theta$)\;
		Backpropagation\;
    $\bm{\theta} \leftarrow \bm{\theta} - \eta \frac{\partial E}{\partial \bm{\theta}}$\;
	}
	Select solution with the best performance\;
	\caption{Gumbel-softmax Optimization (GSO)}
	\label{GSO}
\end{algorithm}

\subsection*{Evolutionary Gumbel-softmax Optimization (EvoGSO)}
In parallelized GSO, simply selecting the result with the best performance from the batch can not take fully advantage of other candidates. Therefore, we propose an improved version of algorithm called Evolutionary Gumbel-softmax Optimization (EvoGSO) by combining evolutionary operators and Gumbel-softmax optimization method. The key idea is to treat a batch as a population so that we can perform population based evolution strategies \cite{Yildiz2012} to improve this algorithm.

Evolution strategy and evolution programming \cite{Back} have shown their capability of solving many optimization problems, they bring diversity to the population and can potentially overcome the difficulty of local minima. In this work, we introduce two types of simple but effective operations to our original GSO algorithm: selective substitution inspired by swarm intelligence and evolutionary operators from genetic algorithm including selection, crossover and mutation.

\subsubsection*{Selective substitution}
During the process of gradient descent, we replace the parameters of worst $1/u$ individuals with a series of alternative parameters every $T_1$ steps. Where, the ratio of substitution $1/u$ and the evolution cycle $T_1$ are hyper-parameters which are varying according to specific problems. The alternative parameters can be the parameters with the best performance in the population, or the best ones with stochastic disturbance, or the ones randomly re-initialized in the problem domain \cite{Back}. This operation is particularly effective on population with high deviation and problems with severe local minima.

\subsubsection*{Selection, crossover and mutation}
When GSO reaches convergence where further optimized solutions cannot be found, we introduce these operators from the classic genetic algorithm to the population for the purpose of diversity and preservation of excellent genes (certain bits or segments of parameters). Here we adopt roulette wheel selection, single-point crossover and binary mutation as well as elitist preservation strategy \cite{davis1991handbook}. Since this operation significantly change the structure of parameters which works against gradient descent, the good performance can be achieved if the evolution operators are implemented after each convergence and with cycle $T_2$ long enough for the population to converge.

We present our proposed method in Algorithm~(\ref{EvoGSO}).

\begin{algorithm}[!htbp]
  \KwIn{Problem size $N$, batch size $N_{\text{bs}}$, learning rate $\eta$, and Graph $\mathcal{G}$ for optimization. Evolution cycle $T$, substitution ratio $1/u$, mutation rate $m$.}
	\KwOut{solution with the best performance}
  Initialize $\bm \theta = (\theta_1, \theta_2, \cdots, \theta_N) \in \mathbb{R}^{N_{bs}\times N \times K}$\;
	\Repeat{Convergence}
	{
		$\mathbf{s} \leftarrow \text{Gumbel-softmax sampling from } p_{\bm \theta} (p_{\bm \theta}=\text{Sigmoid}(\bm \theta))$\;
		$E \leftarrow E(\mathbf s;\bm \theta$)\;
		Backpropagation\;
    $\bm{\theta} \leftarrow \bm{\theta} - \eta \frac{\partial E}{\partial \bm{\theta}}$\;
    \Do{Every $T_1$ steps and the variance of populations is larger than the threshold}{
      select best $1/u$ solutions and worst $1/u$ solutions\;
      replace the parameters of the worst solutions by the parameters of the best solutions\;
    }
    \Do{Every $T_2$ steps after the first convergence of the gradient based steps}{
      retain elite individuals\;
      perform crossover and mutation operation and replace parents\;
    }
	}
	Select solution with the best performance\;
	\caption{EvoGSO}
	\label{EvoGSO}
\end{algorithm}

\section*{Experiments}
\label{sec:Experiments}
\subsection*{A Simple Example}
\begin{figure}[!htbp]
    \centering
    \subfigure[Rastrigin function]{
    \includegraphics[width=0.45\textwidth]{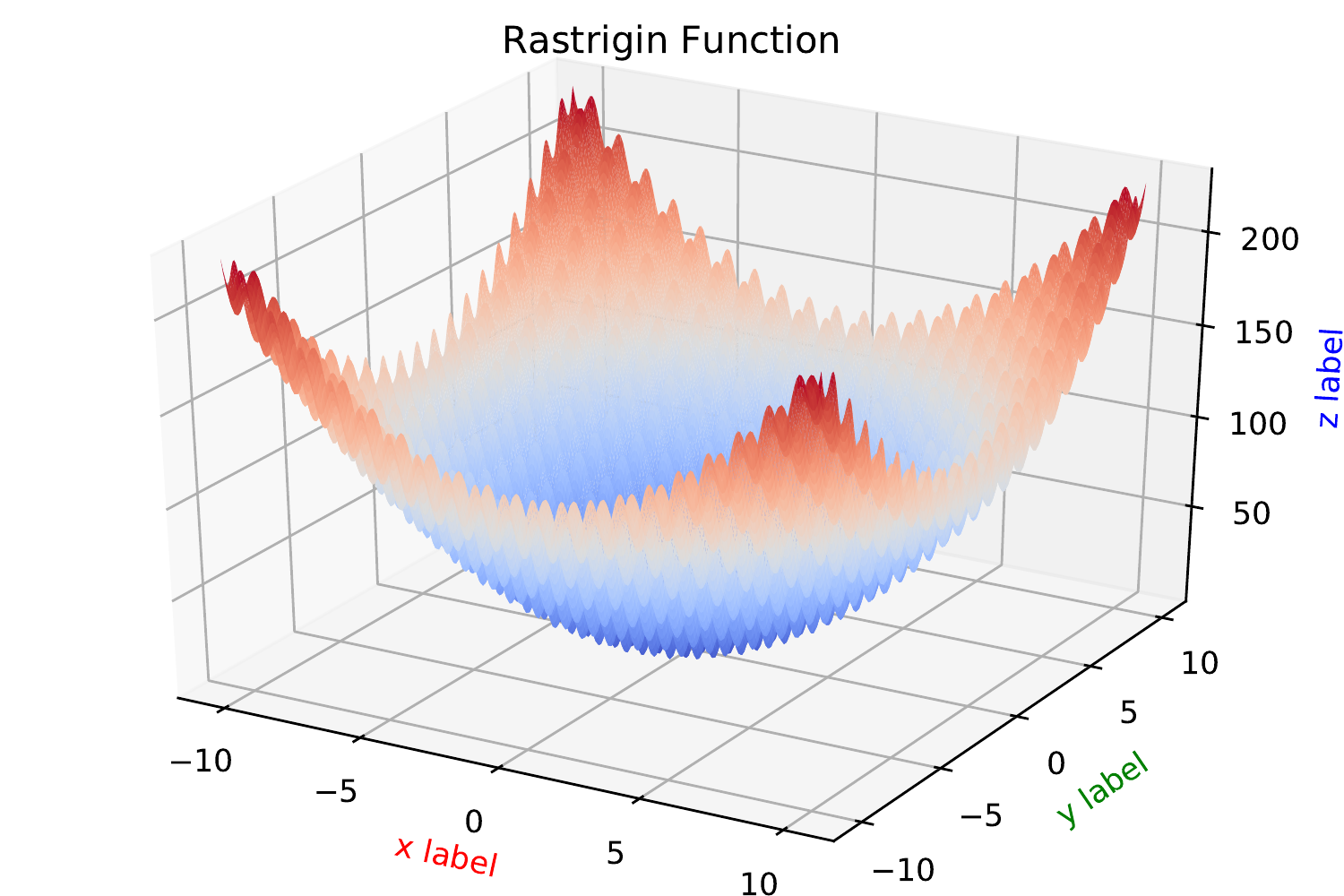} 
}
    \subfigure[Griewank function]{
    \includegraphics[width=0.45\textwidth]{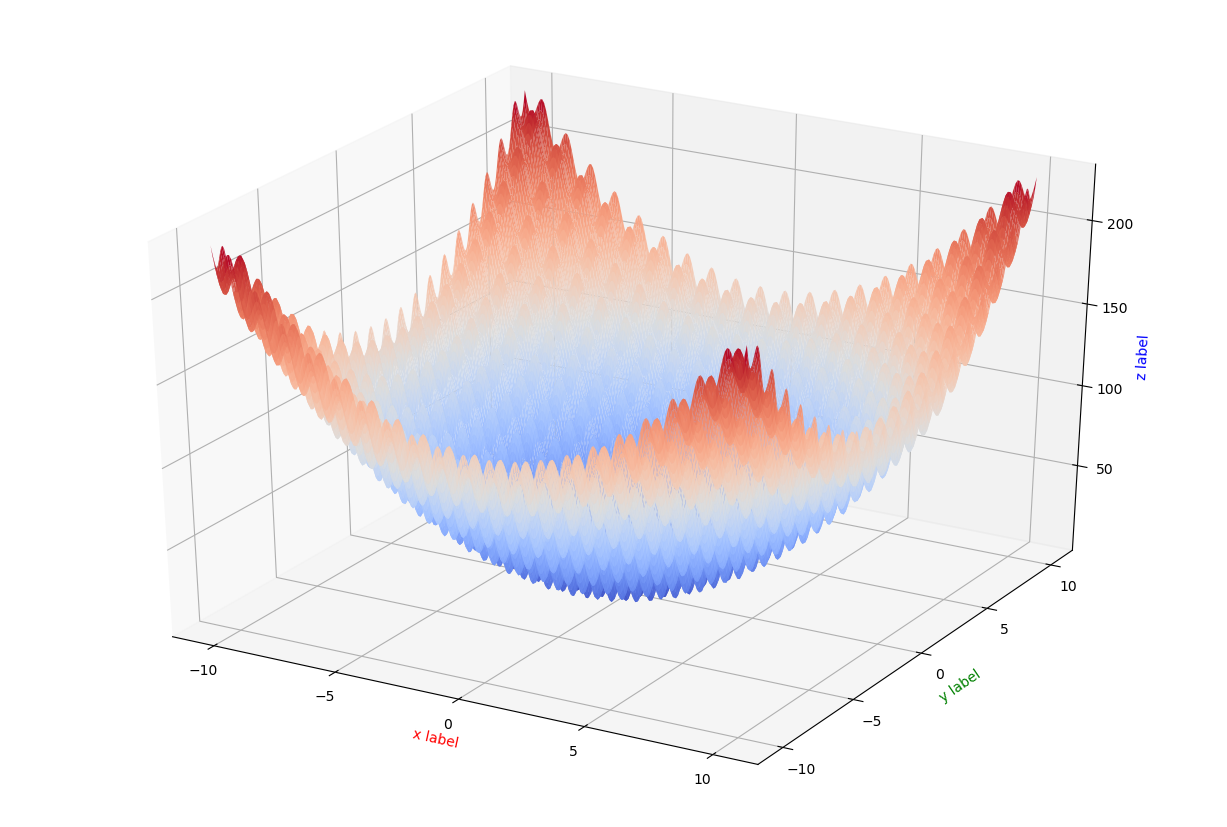}
}
    \caption{Images of two test functions}
    \label{fig:functions}
\end{figure}

To show the importance and the efficiency of combining evolutionary operators and gradient based optimization method, we use a functional optimization problem as an example at first. We test the hybrid algorithm of evolutionary method and gradient based method on functional optimization problem for Griewank and Rastrigin functions (Figure \ref{fig:functions}). These functions are classic test functions for optimization algorithms since they contain lots of local minima, and the global minimum can be hard to find. 

\begin{figure}[!htbp]
    \centering
    \includegraphics[width=0.96\textwidth]{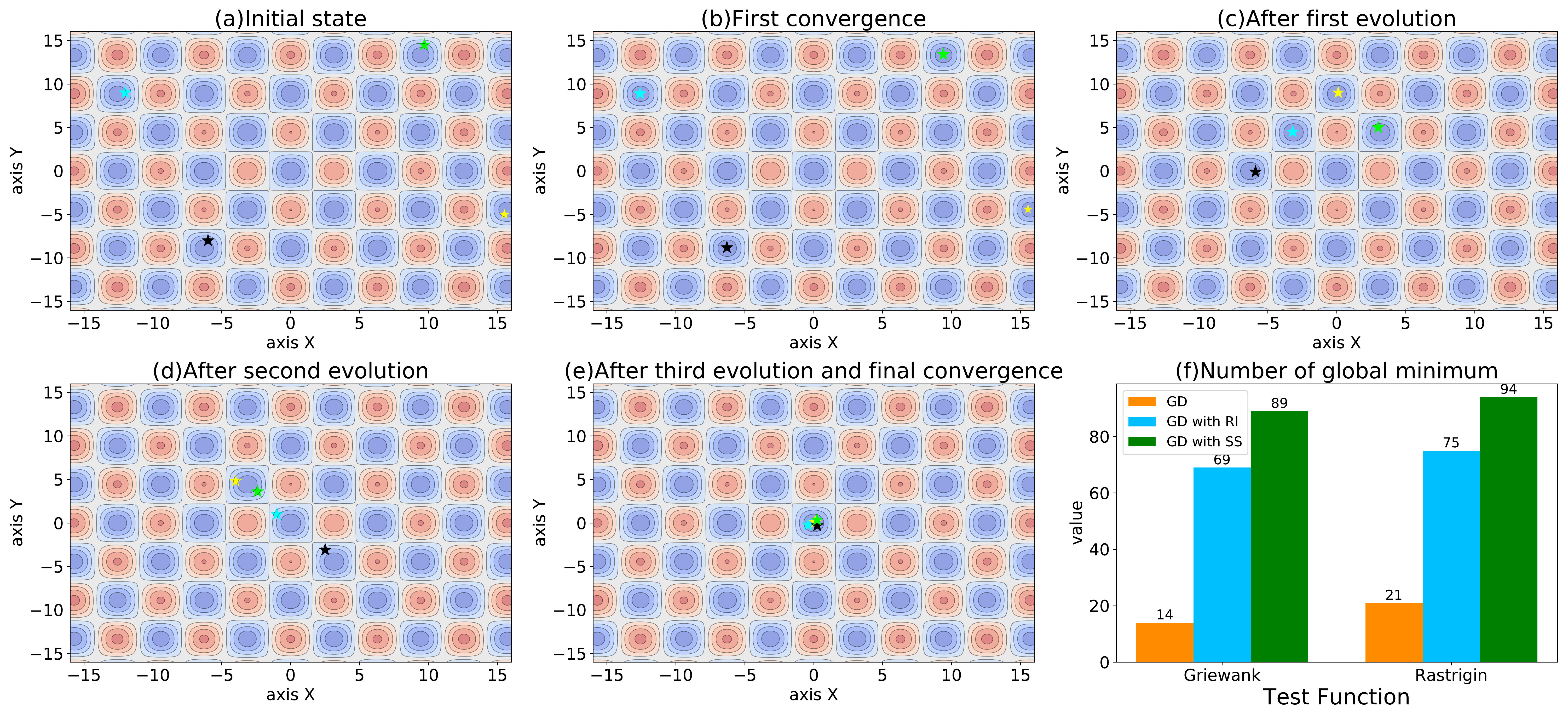}

    \caption{(a) to (e) are five key frames that illustrate how four candidate individuals with different colors converge to the global minimum at $(0,0)$ under the hybrid algorithm on the contour of Griewank function. 
    (a) The initial positions of the four candidates.
    (b) The positions of the four candidates after the first convergence of gradient decent but without evolutionary operation.
    (c) The positions of the four candidates after the first evolutionary operation.
    (d) The positions of the four candidates after the second evolutionary operation.
    (e) The final positions of the four candidates.
    The bar graph in (f) shows the number of global minimums found by GD, GD with random initialization, and GD with selective substitution algorithms in 100 instances, respectively.}
    \label{fig:evolution}
\end{figure}

We run three different optimization algorithms on these functions: gradient descent(GD) with learning rate $\eta$ = 0.01, GD with random initialization with cycle $T$ = 1000 and hybrid algorithm of GD and evolution strategy with population size $N_{\text{bs}}$ = 64, evolution cycle $T$ = 1000 and the substitution ratio $1/u$ = 1/4 (see Figure \ref{fig:evolution} (a)). In gradient descent algorithm, candidates usually stuck in local minima after convergence (see Figure \ref{fig:evolution} (b)). After we add random initialization operation, candidates are able to jump out of these local minima and have more chance to find global minimum(see Figure \ref{fig:evolution} (c) and (d)). However, it is stochastic and candidates are unable to share information with each other. Finally we test a hybrid algorithm of GD and evolution strategy. We adopt selective substitution operation inspired by swarm intelligence, in which candidates are able to communicate so that the good results can be preserved and inherited(see Figure \ref{fig:evolution} (e)). Figure \ref{fig:evolution} illustrate five key frames on contour of Griewank function during the optimizing process of this hybrid algorithm and a comparison bar graph shows the number of global minimum found by different optimization algorithms in 100 instances. We can clearly see that the hybrid algorithm outperforms its two competitors and obtain global minimum more likely.

\subsection*{Combinatorial Optimization Problems on Graphs}
To further test the performance of our proposed algorithms, we conduct experiments on different optimization problems on graphs. We perform all experiments on a server with an Intel Xeon Gold 5218 CPU and NVIDIA GeForce RTX 2080Ti GPUs. 
For comparison, we mainly test the three general optimization algorithms: extremal optimization (EO) \cite{boettcher2000nature}, simulated annealing (SA) \cite{kirkpatrick1983optimization} and genetic algorithm (GA).

\subsubsection*{Modularity maximization}
Modularity is a graph clustering index for detecting community structure in complex networks \cite{fortunato2010community}. Suppose a graph $\mathcal{G(V,E)}$ is partitioned into $K$ communities, the objective is to maximize the following modularity function such that the best partition for nodes can be found,
\begin{equation}
    E(s_1, s_2, \cdots, s_N) = \frac{1}{2\abs{\mathcal{E}}}\sum_{ij}\left[A_{ij}-\frac{k_{i} k_{j}}{2 \abs{\mathcal{E}}}\right]\delta(s_i, s_j),
    \label{eq:modularity}
\end{equation}
where $\abs{\mathcal{E}}$ is the number of edges, $k_i$ is the degree of node $i$, $s_i\in \{0,1,\cdots,K-1\}$ is a label denoting which community of node $i$ belongs to, $\delta(s_i,s_j)=1$ if $s_i=s_j$ and $0$ otherwise. $A_{ij}$ is the adjacent matrix of the graph. Maximizing modularity in general graphs is an NP-hard problem \cite{brandes2007finding}.

We use the real-world datasets that have been well studied in \cite{newman2004fast,newman2006modularity,duch2005community} : \emph{Karate, Jazz, C.elegans and E-mail} to test the algorithms. We run experiments on each dataset with the number of communities $Ncoms$ ranging from 2 to 20. We run 10 instances for each fixed $Ncoms$. After the optimization process for the modularity in all $Ncoms$ values, we report the maximum modularity value $Q$ and the corresponding $Ncoms$ in Table~\ref{tab:mod}. Our proposed methods have achieved competitive modularity values on all datasets compared to hierarchical agglomeration \cite{newman2004fast} and EO \cite{duch2005community}.

\begin{figure}[!htbp]
    \centering
    \includegraphics[width=0.75\linewidth]{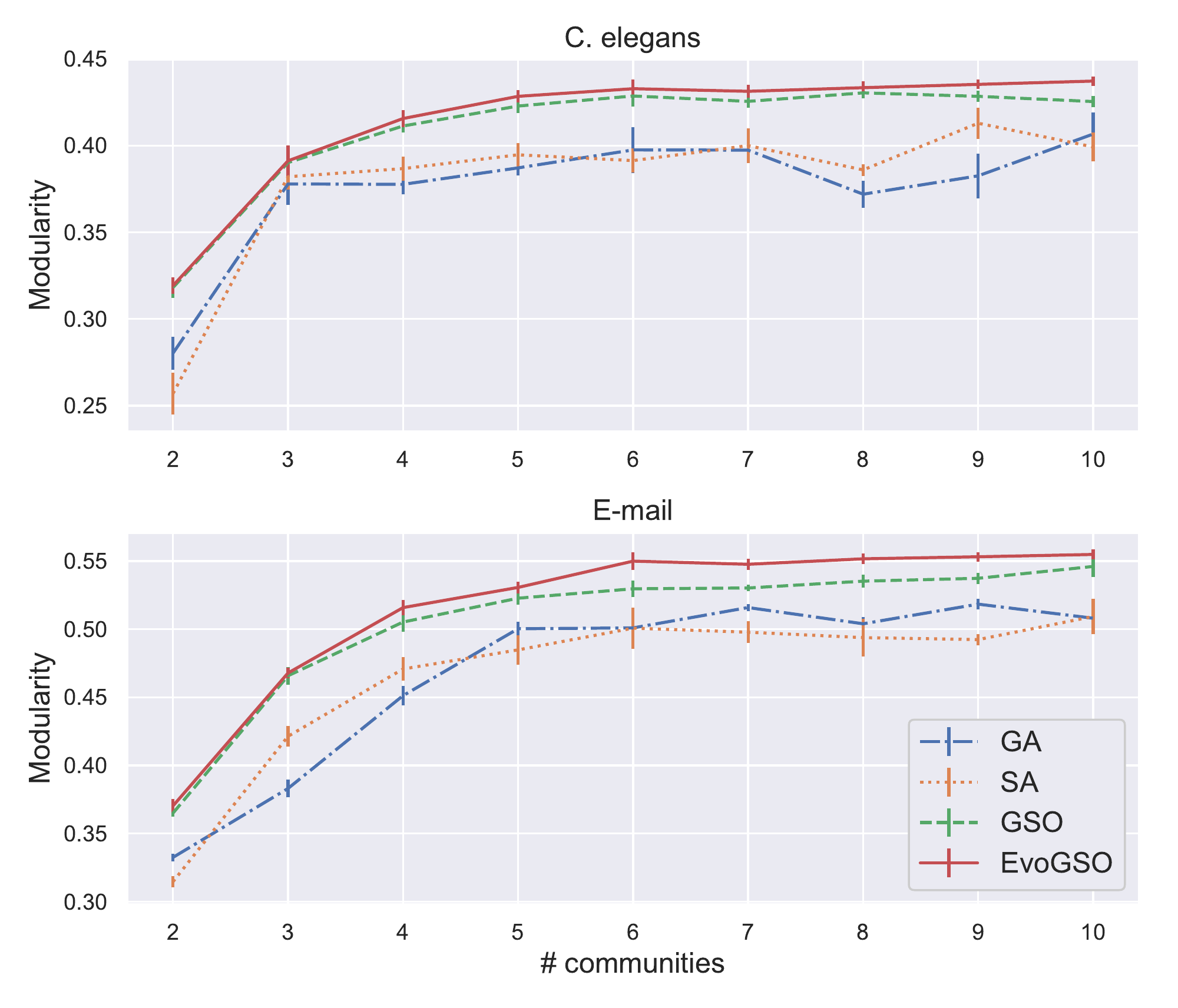}
    \caption{Results on modularity optimization. In experiments, we suppose that the graph is partitioned into $K$ communities with $K$ ranging from 2 to 10 and report the maximum modularity value Q. We only perform experiments on two larger graphs: \emph{C.elegans} and \emph{E-mail} since the sizes of \emph{karate} and \emph{Jazz} are too small. Experiment configuration: (GSO/EvoGSO) : batch size  = 256, initial $\tau$ = 0.5, final $\tau$ = 0.1, learning rate = 0.01, instance = 10, cycle $T_1$ = 100, cycle $T_2$ = 5000, substitution ratio $1/u$ = 1/8, mutation rate $m$ = 0.001, elite ratio = 0.0625. (GA) : population size = 64, crossover rate = 0.8, mutation rate = 0.001, elite ratio = 0.125.}
    \label{fig:mod}
\end{figure}

\begin{table}[!htbp]
  \centering
  \begin{threeparttable}
  \caption{Results on modularity optimization.\tnote{1,2}}
  \label{tab:mod}
  \begin{tabular}{@{}cccccc@{}}
  \toprule
  Graph     & Size & \cite{newman2004fast}   & EO \cite{duch2005community}        & GSO\tnote{3}      & EvoGSO\tnote{4}    \\ \midrule
  Karate    & 34   & 0.3810/2  & \mystar{0.4188}/4  & \textbf{0.4198}/4 & \textbf{0.4198}/4  \\
  Jazz      & 198  & 0.4379/4  & \textbf{0.4452}/5  & \mystar{0.4451}/4 & \mystar{0.4451}/4  \\
  C.elegans & 453  & 0.4001/10 & \mystar{0.4342}/12 & 0.4304/8 & \textbf{0.4418}/11 \\
  E-mail    & 1133 & 0.4796/13 & \textbf{0.5738}/15 & 0.5275/8 & \mystar{0.5655}/15 \\ \bottomrule
  \end{tabular}
      \begin{tablenotes}
       \footnotesize
       \item[1] We report the maximum modularity value $Q$ and the corresponding number of communities $Ncoms$ in the form of ($Q$/$Ncoms$).
       \item[2] The best and the second best results are denoted in bold and asterisk respectively.
       \item[3] Configuration: batch size  = 256, initial $\tau$ = 0.5, final $\tau$ = 0.1, learning rate = 0.01, instance = 10.
       \item[4] Configuration: batch size = 256, initial $\tau$ = 0.5, final $\tau$ = 0.1, learning rate = 0.01, instance = 10, cycle $T_1$ = 100, cycle $T_2$ = 5000, substitution ratio $1/u$ = 1/8, mutation rate $m$ = 0.001, elite ratio = 0.0625.
     \end{tablenotes}
     \end{threeparttable}
\end{table}
Figure~\ref{fig:mod} further shows the modularity value with different number of communities on \emph{C.elegans} and \emph{E-mail}. Comparing to GA and SA, our proposed methods have achieved much higher modularity for different number of communities.

\subsubsection*{Sherrington-Kirkpatrick (SK) model}
SK model is a celebrated spin glasses model defined on a complete graph \cite{sherrington1975solvable}. Each node represents an Ising spin $\sigma_i \in \{-1, +1\}$ and the interaction between spins $\sigma_i$ and $\sigma_j$ is $J_{ij}$ sampled from a Gaussian distribution $\mathcal{N}(0, 1/N)$ where $N$ is the number of spins. We are asked to give an assignment of each spin so that the objective function, or the ground state energy
\begin{equation}
    E(\sigma_1, \sigma_2, \cdots, \sigma_N) = -\sum_{1 \leq i < j \leq N} J_{ij} \sigma_i \sigma_j
    \label{eq:sk_model}
\end{equation}
is minimized. It is also an NP-hard problem \cite{mezard1987spin}.

We test our algorithms on SK model with various sizes ranging from 256 to 8192. The state-of-the-art results are obtained by EO \cite{boettcher2000nature}. The results are shown in Table~\ref{tab:sk-1} and Table~\ref{tab:sk-2}.
From Table~\ref{tab:sk-1} we see that although EO has obtained lower ground state energy, it only reported results of system size up to $N=1024$ because it is extremely time-consuming. In fact, the algorithmic cost of EO is $\mathcal{O}(N^4)$. 
In the implementation of SA and GA, we set the time limit to be 96 hours and the program failed to finish for some $N$ in both algorithms. Although the results of SA are much better than GA, they are still not satisfying for larger $N$. 
For SK model, we adopt only selective substitution in EvoGSO. 

\begin{table}[!htbp]
\centering
\begin{threeparttable}
\caption{The results on optimization of ground state energy of SK model compared to Extremal optimization (EO), genetic algorithm (GA) and simulated annealing (SA).\tnote{1}}
\label{tab:sk-1}
\begin{tabular}{@{}cccccc@{}}
\toprule
N    & I    & EO \cite{boettcher2005extremal}                     & GA\tnote{2}                 & SA                   & GSO ($N_{bs}=1$)\tnote{3}        \\ \midrule
256  & 5000 & \textbf{-0.74585(2)}/$\sim$268s & -0.6800(3)/16.3s   & -0.7278(2)/1.28s     & -0.7267(2)/0.99s  \\
512  & 2500 & \textbf{-0.75235(3)}/$\sim$1.2h & -0.6580(3)/60.06s  & -0.7327(2)/3.20s     & -0.7405(2)/2.16s  \\
1024 & 1250 & \textbf{0.7563(2)}/$\sim$20h    & -0.6884(4)/236.21s & -0.7352(2)/15.27s    & -0.7480(2)/4.49s  \\
2048 & 400  & -                      & -                  & -0.7367(2)/63.27s    & \textbf{-0.7524(2)}/7.23s  \\
4096 & 200  & -                      & -                  & -0.73713(6)/1591.93s & \textbf{-0.7551(2)}/10.46s \\
8192 & 100  & -                      & -                  & -                    & \textbf{-0.7562(1)}/25.15s \\ \bottomrule
\end{tabular}
    \begin{tablenotes}
       \footnotesize
        \item[1] The best results are denoted in bold. Corresponding standard error of the mean is given in brackets.
        \item[2] Configuration: population size = 64, crossover rate = 0.8, mutation rate = 0.001, elite ratio = 0.125.
        \item[3] Configuration: initial $\tau$ = 20, final $\tau$ = 1, learning rate = 1.
     \end{tablenotes}
     \end{threeparttable}
\end{table}

We also compare Gumbel-softmax based algorithms with different batch sizes and the EvoGSO. From Table~\ref{tab:sk-2} we see that with the implementation of the parallel version, the results can be improved greatly. Besides, the EvoGSO outperforms GSO for larger $N$.

\begin{table}[!htbp]
\centering
\begin{threeparttable}
\caption{The results on optimization of ground state energy of SK model. We show that the parallel version of our proposed methods and EvoGSO can greatly improve the performance.\tnote{1}}
\label{tab:sk-2}
\begin{tabular}{@{}ccccc@{}}
\toprule
N    & I    & GSO ($N_{bs}=1$)\tnote{2}        & GSO ($N_{bs}=128$)\tnote{2}      & EvoGSO ($N_{bs}=128$)\tnote{3}   \\ \midrule
256  & 5000 & -0.7267(2)/0.99s  & \textbf{-0.7369(1)}/0.96s  & -0.7364(1)/0.89s  \\
512  & 2500 & -0.7405(2)/2.16s  & \textbf{-0.7464(1)}/2.14s  & -0.7462(1)/2.01s  \\
1024 & 1250 & -0.7480(2)/4.49s  & \textbf{-0.7521(1)}/4.66s  & -0.7516(4)/4.41s  \\
2048 & 400  & -0.7524(2)/7.23s  & -0.7555(2)/8.07s  & \textbf{-0.7557(1)}/7.51s  \\
4096 & 200  & -0.7551(2)/10.46s & -0.7566(5)/12.78s & \textbf{-0.7569(3)}/12.80s \\
8192 & 100  & -0.7562(1)/25.15s & -0.7568(8)/49.13s & \textbf{-0.7578(5)}/49.04s \\ \bottomrule
\end{tabular}
    \begin{tablenotes}
       \footnotesize
        \item[1] The best results are denoted in bold. The corresponding standard error of the mean is given in brackets.
        \item[2] Configuration: initial $\tau$ = 20, final $\tau$ = 1, learning rate = 1.
        \item[3] Configuration: initial $\tau$ = 20, final $\tau$ = 1, learning rate = 1, cycle $T_1$ = 100, substitution ratio $1/u$ = 1/8.
     \end{tablenotes}
     \end{threeparttable}
\end{table}

\subsubsection*{Maximal Independent Set (MIS) and Minimum Vertex Cover (MVC) problems}
MIS and MVC problems are canonical NP-hard combinatorial optimization problems on graphs \cite{karp1972reducibility}. Given an undirected graph $\mathcal{G(V,E)}$, the MIS problem asks to find the largest subset $\mathcal V^{\prime} \subseteq \mathcal V$ such that no two nodes in $\mathcal V^{\prime}$ are connected by an edge in $\mathcal E$. Similarly, the MVC problem asks to find the smallest subset $\mathcal V^{\prime} \subseteq \mathcal V$ such that every edge in $\mathcal{E}$ is incident to a node in $\mathcal{V^{\prime}}$. MIS and MVC are constrained optimization problems and cannot be optimized directly by our framework. Here we adopt penalty method and Ising formulation to transform them into unconstrained problems.

We can place an Ising spin $\sigma_i$ on each node and then define the binary bit variable $x_i = (\sigma_i + 1)/2$. Here $x_i = 1$ means that node $i$ belongs to the subset $\mathcal{V^{\prime}}$ and $x_i=0$ otherwise. Thus the Ising Hamiltonians for MIS problem is
\begin{equation}
    E(x_1, x_2, \cdots, x_N) = -\sum_i x_i + \alpha \sum_{ij\in \mathcal E}x_i x_j,
    \label{eq:mis}
\end{equation}
Similarly, the Ising Hamiltonians for MVC becomes
\begin{equation}
    E(x_1, x_2, \cdots, x_N) = \sum_i x_i + \alpha \sum_{ij \in \mathcal{E}} (1-x_i)(1-x_j).
    \label{eq:mvc}
\end{equation}
where $\alpha > 0$.
The first term on right hand side is the number of selected nodes and the second term provides a penalty if selected nodes violate constraint. $\alpha$ is a penalty parameter and its value is crucial to the performance of our framework. If $\alpha$ is set too small, we may not find any feasible solutions. Conversely, if it is set too big, we may find lots of feasible solutions whose qualities are not satisfying. In this work, we set $\alpha$ to $3$, which assures both quality and amount of feasible solutions.

We test our algorithms on three citation graphs: \emph{Cora, Citeseer and PubMed}. Beyond the standard general algorithms like Genetic Algorithm and Simulating Anealing, we also compare with other deep learning based algorithms including (1) Structure2Vec Deep Q-learning (S2V-DQN) \cite{khalil2017learning}: a reinforcement learning method to address optimization problems over graphs, and (2) Graph Convolutional Networks with Guided Tree Search (GCNGTS) \cite{li2018combinatorial}: a supervised learning method based on graph convolutional networks (GCN) \cite{kipf2016semi}, as well as the well known greedy algorithms on MIS and MVC problems like (3) greedy algorithm (Greedy) and Minimum-degree greedy algorithm (MD-Greedy) \cite{halldorsson1997greed}: a simple and well-studied method for finding independent sets in graphs. 

We run 20 instances and report results with best performance.
The results of MIS and MVC problems are shown in Table~\ref{tab:mis-mvc}.
Our proposed algorithms have obtained much better results compared to the classical general optimization methods including greedy and SA on all three datasets. Although our methods cannot beat MD-Greedy algorithm, they do not use any prior information about the graph. However, MD-Greedy requires to compute degrees of all nodes on the graph. Further, we do not report the results of GA algorithm because without heuristic and specific design, the general GA failed to find any feasible solution since MIS and MVC are constrained optimization problems.

\begin{table}[!htbp]
\centering
\begin{threeparttable}
\caption{Results on MIS and MVC problems compared to classic methods and supervised deep learning methods.\tnote{1}}
\label{tab:mis-mvc}
\begin{tabular}{@{}ccc|ccc|cc|c@{}}
\toprule
                     & \multicolumn{2}{c}{Graph Info} & \multicolumn{3}{c}{Classic} & \multicolumn{2}{c}{Supervised} & Proposed   \\
                     & Name            & Size         & MD-Greedy  & Greedy & SA    & S2V-DQN        & GCNGTS        & GSO\tnote{2} / EvoGSO\tnote{3} \\ \midrule
\multirow{3}{*}{MIS} & Cora            & 2708         & \textbf{1451}    & 672       & 1390  & 1381           & \textbf{1451}          & \mystar{1443}       \\
                     & Citeseer        & 3327         & \mystar{1818}    & 1019      & 1728  & 1705           & \textbf{1867}          & 1795       \\
                     & PubMed          & 19717        & \textbf{15912}   & 5353      & 14703 & 15709          & \textbf{15912}         & \mystar{15886}      \\ \midrule
\multirow{3}{*}{MVC} & Cora            & 2708         & \textbf{1257}    & 2036      & 1318  & 1327           & \textbf{1257}          & \mystar{1265}       \\
                     & Citeseer        & 3327         & \mystar{1509}    & 2308      & 1599  & 1622           & \textbf{1460}          & 1533       \\
                     & PubMed          & 19717        & \textbf{3805}    & 14364     & 5014  & 4008           & \textbf{3805}          & \mystar{3831}       \\ \bottomrule
\end{tabular}
    \begin{tablenotes}
       \footnotesize
       \item[1] The best and the second best results are denoted in bold and asterisk respectively.
       \item[2] Configuration: batch size = 128, fixed $\tau$ = 1, learning rate = 0.01, $\alpha$ = 3, instance = 20.
       \item[3] Configuration: batch size = 512, fixed $\tau$ = 1, learning rate = 0.01, $\alpha$ = 3, instance = 20, cycle $T_1$ = 100, substitution ratio $1/u$ = 1/8, cycle $T_2$ = 10000, mutation rate $m$ = 0.001, elite ratio = 0.0625. 
     \end{tablenotes}
     \end{threeparttable}
\end{table}

It is necessary to emphasize that the differences between our framework and other deep learning based algorithms such as S2V-DQN and GCNGTS. These algorithms belong to supervised learning, thus contain two stages of problem solving: training the solver at first, and then testing. Although relatively good solutions can be obtained efficiently, they must consume a great deal of time for training the solver and the quality of solutions depend heavily on the quality and the amount of the data for training. These features can hardly extend for large graphs. Comparatively, our proposed framework is more direct and light-weight, for it contains only optimization stage. It requires no training part and has no dependence on data or specific domain knowledge at all, therefore can easily be generalized and modified for different optimization problems.

\subsubsection*{Sensitivity analysis on hyper-parameters}
\begin{figure}[!htbp]
    \centering
    \includegraphics[width=0.95\linewidth]{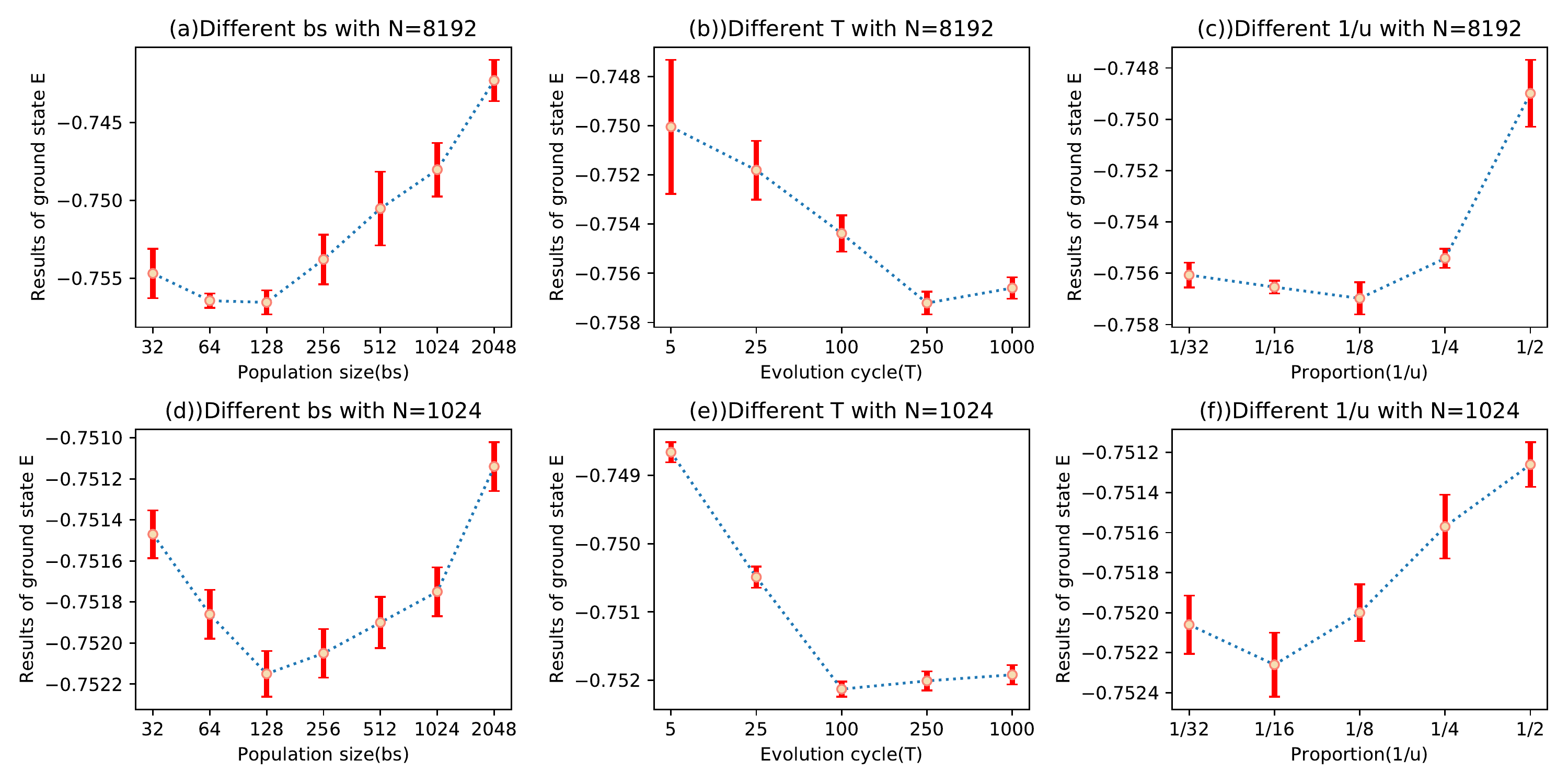}
    \caption{Results on hyper-parameters tuning of population size $N_{bs}$, evolution cycle $T$ and substitution ratio $1/u$ on SK model with 1024 and 8192 nodes. Experiment configuration: initial $\tau = 20$, final $\tau = 1$, learning rate $\eta$ = 1. The results are averaged for 1250 instances with 1024 nodes and 100 instances with 8192 nodes respectively.}
    \label{fig:hp}
\end{figure}

We also perform experiments to test how hyper-parameters in evolution operation affects the performance of our algorithms. We have tried different population size $N_{bs}$, evolution cycle $T_1$ and substitution ratio $1/u$ on SK model with 1024 and 8192 nodes. The default configurations are: initial $\tau = 20$, final $\tau = 1$, learning rate $\eta$ = 1, $N_{bs} = 128$, $T_1 = 100$, $1/u = 1/8$, and then we change one hyper-parameter every time for test. The results are shown in Figure~\ref{fig:hp} . We can see that our framework shows different sensitivity to these hyper-parameters as they changes, and a relatively satisfying combination of hyper-parameters can be given from this research.

\section*{Conclusion}
\label{sec:Conclusion}
In this work, we present a simple general framework for solving optimization problems on graphs. Our method is based on advanced automatic differentiation techniques and Gumbel-softmax technique which allows the gradients passing through sampling processes directly. We assume that all nodes in the network are independent and thus the joint distribution is factorized as a product distributions of each node. This enables Gumbel-softmax sampling process efficiently. Furthermore, we introduce evolution strategy into our framework, which brings diversity and improves the performance of our algorithm. Our experiment results show that our method has good performance on all three tasks and also take advantages in time complexity. Comparing to the traditional general optimization methods such as GA and SA, our framework can tackle large graphs easily and efficiently. Though not competitive to state-of-the-art deep learning based method, our framework has the advantage of requiring neither training the solver nor specific domain knowledge. In general, it is an efficient, general and lightweight optimization framework for solving optimization problems on graphs. 

However, there is much space to improve our algorithm on accuracy. In this paper, we take the mean field approximation as our basic assumption, however, the variables are not independent on most problems. Therefore, much more sophisticated variational distributions can be considered in the future. Another way to improve accuracy is to combine other skills such as local search in our framework. Since our framework is general and requires no specific domain knowledge, it shall be tested for solving other complex optimization problems in the future.


\begin{backmatter}
\section*{Abbreviations}
GSO: Gumbel-softmax optimization; EvoGSO: Evolutionary Gumbel-softmax optimization.

\section*{Acknowledgements}
This research is supported by the National Natural Science Foundation of China (NSFC) (no. 61673070) and the Fundamental Research Funds for the Central Universities (no. 2020KJZX004).

\section*{Author's contributions}
J.Z., Y.L.and J.L. conceived and designed the research. Y.L., J.L.and J.Z. designed the model structure. Y.L. and J.L. developed the model. Y.L., J.L., G.L., Y.H.and M.M. performed the experiments. J.Z., Y.L. and J.L. wrote the manuscript. J.Z.reviewed and revised the manuscript. J.Z.supervised the research.

\section*{Funding}
Not applicable.

\section*{Availability of data and materials}
The dataset analyzed in this study is publicly available online at \url{http://networkrepository.com/}.

\section*{Competing interests}
  The authors declare that they have no competing interests.


\bibliographystyle{bmc-mathphys} 
\bibliography{bmc_article}      


\begin{thebibliography}{31}
\ifx \bisbn   \undefined \def \bisbn  #1{ISBN #1}\fi
\ifx \binits  \undefined \def \binits#1{#1}\fi
\ifx \bauthor  \undefined \def \bauthor#1{#1}\fi
\ifx \batitle  \undefined \def \batitle#1{#1}\fi
\ifx \bjtitle  \undefined \def \bjtitle#1{#1}\fi
\ifx \bvolume  \undefined \def \bvolume#1{\textbf{#1}}\fi
\ifx \byear  \undefined \def \byear#1{#1}\fi
\ifx \bissue  \undefined \def \bissue#1{#1}\fi
\ifx \bfpage  \undefined \def \bfpage#1{#1}\fi
\ifx \blpage  \undefined \def \blpage #1{#1}\fi
\ifx \burl  \undefined \def \burl#1{\textsf{#1}}\fi
\ifx \doiurl  \undefined \def \doiurl#1{\textsf{#1}}\fi
\ifx \betal  \undefined \def \betal{\textit{et al.}}\fi
\ifx \binstitute  \undefined \def \binstitute#1{#1}\fi
\ifx \binstitutionaled  \undefined \def \binstitutionaled#1{#1}\fi
\ifx \bctitle  \undefined \def \bctitle#1{#1}\fi
\ifx \beditor  \undefined \def \beditor#1{#1}\fi
\ifx \bpublisher  \undefined \def \bpublisher#1{#1}\fi
\ifx \bbtitle  \undefined \def \bbtitle#1{#1}\fi
\ifx \bedition  \undefined \def \bedition#1{#1}\fi
\ifx \bseriesno  \undefined \def \bseriesno#1{#1}\fi
\ifx \blocation  \undefined \def \blocation#1{#1}\fi
\ifx \bsertitle  \undefined \def \bsertitle#1{#1}\fi
\ifx \bsnm \undefined \def \bsnm#1{#1}\fi
\ifx \bsuffix \undefined \def \bsuffix#1{#1}\fi
\ifx \bparticle \undefined \def \bparticle#1{#1}\fi
\ifx \barticle \undefined \def \barticle#1{#1}\fi
\ifx \bconfdate \undefined \def \bconfdate #1{#1}\fi
\ifx \botherref \undefined \def \botherref #1{#1}\fi
\ifx \url \undefined \def \url#1{\textsf{#1}}\fi
\ifx \bchapter \undefined \def \bchapter#1{#1}\fi
\ifx \bbook \undefined \def \bbook#1{#1}\fi
\ifx \bcomment \undefined \def \bcomment#1{#1}\fi
\ifx \oauthor \undefined \def \oauthor#1{#1}\fi
\ifx \citeauthoryear \undefined \def \citeauthoryear#1{#1}\fi
\ifx \endbibitem  \undefined \def \endbibitem {}\fi
\ifx \bconflocation  \undefined \def \bconflocation#1{#1}\fi
\ifx \arxivurl  \undefined \def \arxivurl#1{\textsf{#1}}\fi
\csname PreBibitemsHook\endcsname

\bibitem{karp1972reducibility}
\begin{bchapter}
\bauthor{\bsnm{Karp}, \binits{R.M.}}:
\bctitle{Reducibility among combinatorial problems}.
In: \bbtitle{Complexity of Computer Computations},
pp. \bfpage{85}--\blpage{103}.
\bpublisher{Springer}, \blocation{???}
(\byear{1972})
\end{bchapter}
\endbibitem

\bibitem{mezard1987spin}
\begin{bbook}
\bauthor{\bsnm{M{\'e}zard}, \binits{M.}},
\bauthor{\bsnm{Parisi}, \binits{G.}},
\bauthor{\bsnm{Virasoro}, \binits{M.}}:
\bbtitle{Spin Glass Theory and Beyond: An Introduction to the Replica Method
  and Its Applications}
vol. \bseriesno{9}.
\bpublisher{World Scientific Publishing Company}, \blocation{???}
(\byear{1987})
\end{bbook}
\endbibitem

\bibitem{newman2006modularity}
\begin{barticle}
\bauthor{\bsnm{Newman}, \binits{M.E.}}:
\batitle{Modularity and community structure in networks}.
\bjtitle{Proceedings of the national academy of sciences}
\bvolume{103}(\bissue{23}),
\bfpage{8577}--\blpage{8582}
(\byear{2006})
\end{barticle}
\endbibitem

\bibitem{Wright2015}
\begin{barticle}
\bauthor{\bsnm{Wright}, \binits{S.J.}}:
\batitle{{Coordinate descent algorithms}}.
\bjtitle{Mathematical Programming}
\bvolume{151}(\bissue{1}),
\bfpage{3}--\blpage{34}
(\byear{2015}).
doi:\doiurl{10.1007/s10107-015-0892-3}.
\arxivurl{1502.04759}
\end{barticle}
\endbibitem

\bibitem{kennedy95particle}
\begin{bchapter}
\bauthor{\bsnm{Kennedy}, \binits{J.}},
\bauthor{\bsnm{Eberhart}, \binits{R.C.}}:
\bctitle{Particle swarm optimization}.
In: \bbtitle{Proceedings of the IEEE International Conference on Neural
  Networks},
pp. \bfpage{1942}--\blpage{1948}
(\byear{1995})
\end{bchapter}
\endbibitem

\bibitem{kirkpatrick1983optimization}
\begin{barticle}
\bauthor{\bsnm{Kirkpatrick}, \binits{S.}},
\bauthor{\bsnm{Gelatt}, \binits{C.D.}},
\bauthor{\bsnm{Vecchi}, \binits{M.P.}}:
\batitle{Optimization by simulated annealing}.
\bjtitle{science}
\bvolume{220}(\bissue{4598}),
\bfpage{671}--\blpage{680}
(\byear{1983})
\end{barticle}
\endbibitem

\bibitem{davis1991handbook}
\begin{botherref}
\oauthor{\bsnm{Davis}, \binits{L.}}:
Handbook of genetic algorithms
(1991)
\end{botherref}
\endbibitem

\bibitem{boettcher2000nature}
\begin{barticle}
\bauthor{\bsnm{Boettcher}, \binits{S.}},
\bauthor{\bsnm{Percus}, \binits{A.}}:
\batitle{Nature's way of optimizing}.
\bjtitle{Artificial Intelligence}
\bvolume{119}(\bissue{1-2}),
\bfpage{275}--\blpage{286}
(\byear{2000})
\end{barticle}
\endbibitem

\bibitem{andrade2012fast}
\begin{barticle}
\bauthor{\bsnm{Andrade}, \binits{D.V.}},
\bauthor{\bsnm{Resende}, \binits{M.G.}},
\bauthor{\bsnm{Werneck}, \binits{R.F.}}:
\batitle{Fast local search for the maximum independent set problem}.
\bjtitle{Journal of Heuristics}
\bvolume{18}(\bissue{4}),
\bfpage{525}--\blpage{547}
(\byear{2012})
\end{barticle}
\endbibitem

\bibitem{paszke2017automatic}
\begin{bchapter}
\bauthor{\bsnm{Paszke}, \binits{A.}},
\bauthor{\bsnm{Gross}, \binits{S.}},
\bauthor{\bsnm{Chintala}, \binits{S.}},
\bauthor{\bsnm{Chanan}, \binits{G.}},
\bauthor{\bsnm{Yang}, \binits{E.}},
\bauthor{\bsnm{DeVito}, \binits{Z.}},
\bauthor{\bsnm{Lin}, \binits{Z.}},
\bauthor{\bsnm{Desmaison}, \binits{A.}},
\bauthor{\bsnm{Antiga}, \binits{L.}},
\bauthor{\bsnm{Lerer}, \binits{A.}}:
\bctitle{Automatic differentiation in pytorch}.
In: \bbtitle{NIPS-W}
(\byear{2017})
\end{bchapter}
\endbibitem

\bibitem{williams1992simple}
\begin{barticle}
\bauthor{\bsnm{Williams}, \binits{R.J.}}:
\batitle{Simple statistical gradient-following algorithms for connectionist
  reinforcement learning}.
\bjtitle{Machine learning}
\bvolume{8}(\bissue{3-4}),
\bfpage{229}--\blpage{256}
(\byear{1992})
\end{barticle}
\endbibitem

\bibitem{gumbel}
\begin{bchapter}
\bauthor{\bsnm{Jang}, \binits{E.}},
\bauthor{\bsnm{Gu}, \binits{S.}},
\bauthor{\bsnm{Poole}, \binits{B.}}:
\bctitle{Categorical reparameterization with gumbel-softmax}.
In: \bbtitle{5th International Conference on Learning Representations, {ICLR}
  2017, Toulon, France, April 24-26, 2017, Conference Track Proceedings}.
\bpublisher{OpenReview.net}, \blocation{???}
(\byear{2017}).
\burl{https://openreview.net/forum?id=rkE3y85ee}
\end{bchapter}
\endbibitem

\bibitem{concrete}
\begin{bchapter}
\bauthor{\bsnm{Maddison}, \binits{C.J.}},
\bauthor{\bsnm{Mnih}, \binits{A.}},
\bauthor{\bsnm{Teh}, \binits{Y.W.}}:
\bctitle{The concrete distribution: {A} continuous relaxation of discrete
  random variables}.
In: \bbtitle{5th International Conference on Learning Representations, {ICLR}
  2017, Toulon, France, April 24-26, 2017, Conference Track Proceedings}
(\byear{2017}).
\burl{https://openreview.net/forum?id=S1jE5L5gl}
\end{bchapter}
\endbibitem

\bibitem{Andreasson2007}
\begin{botherref}
\oauthor{\bsnm{Andreasson}, \binits{N.}},
\oauthor{\bsnm{Evgrafov}, \binits{A.}},
\oauthor{\bsnm{Patriksson}, \binits{M.}}:
{An Introduction to Continuous Optimization: Foundations and Fundamental
  Algorithms},
400
(2007)
\end{botherref}
\endbibitem

\bibitem{Avraamidou2020}
\begin{bbook}
\bauthor{\bsnm{Avraamidou}, \binits{S.}},
\bauthor{\bsnm{Pistikopoulos}, \binits{E.N.}}:
\bbtitle{{Optimization of Complex Systems: Theory, Models, Algorithms and
  Applications}}
vol. \bseriesno{991},
pp. \bfpage{579}--\blpage{588}.
\bpublisher{Springer}, \blocation{???}
(\byear{2020}).
doi:\doiurl{10.1007/978-3-030-21803-4}.
\burl{http://link.springer.com/10.1007/978-3-030-21803-4}
\end{bbook}
\endbibitem

\bibitem{Zidani2020}
\begin{bchapter}
\bauthor{\bsnm{Zidani}, \binits{H.}},
\bauthor{\bsnm{Ellaia}, \binits{R.}},
\bauthor{\bparticle{de} \bsnm{Cursi}, \binits{E.S.}}:
\bctitle{{A Hybrid Simplex Search for Global Optimization with Representation
  Formula and Genetic Algorithm}}.
In: \bbtitle{Advances in Intelligent Systems and Computing},
vol. \bseriesno{991},
pp. \bfpage{3}--\blpage{15}.
\bpublisher{Springer}, \blocation{???}
(\byear{2020})
\end{bchapter}
\endbibitem

\bibitem{Rocha2020}
\begin{bchapter}
\bauthor{\bsnm{Rocha}, \binits{A.M.A.}},
\bauthor{\bsnm{Costa}, \binits{M.F.P.}},
\bauthor{\bsnm{Fernandes}, \binits{E.M.}}:
\bctitle{A population-based stochastic coordinate descent method}.
In: \bbtitle{World Congress on Global Optimization},
pp. \bfpage{16}--\blpage{25}
(\byear{2019}).
\bcomment{Springer}
\end{bchapter}
\endbibitem

\bibitem{Yildiz2012}
\begin{barticle}
\bauthor{\bsnm{Yildiz}, \binits{A.R.}}:
\batitle{{A comparative study of population-based optimization algorithms for
  turning operations}}.
\bjtitle{Information Sciences}
\bvolume{210},
\bfpage{81}--\blpage{88}
(\byear{2012}).
doi:\doiurl{10.1016/j.ins.2012.03.005}
\end{barticle}
\endbibitem

\bibitem{liu2019gumbel}
\begin{bchapter}
\bauthor{\bsnm{Liu}, \binits{J.}},
\bauthor{\bsnm{Gao}, \binits{F.}},
\bauthor{\bsnm{Zhang}, \binits{J.}}:
\bctitle{Gumbel-softmax optimization: A simple general framework for
  combinatorial optimization problems on graphs}.
In: \bbtitle{International Conference on Complex Networks and Their
  Applications},
pp. \bfpage{879}--\blpage{890}
(\byear{2019}).
\bcomment{Springer}
\end{bchapter}
\endbibitem

\bibitem{wainwright2008graphical}
\begin{barticle}
\bauthor{\bsnm{Wainwright}, \binits{M.J.}},
\bauthor{\bsnm{Jordan}, \binits{M.I.}}, \betal:
\batitle{Graphical models, exponential families, and variational inference}.
\bjtitle{Foundations and Trends{\textregistered} in Machine Learning}
\bvolume{1}(\bissue{1--2}),
\bfpage{1}--\blpage{305}
(\byear{2008})
\end{barticle}
\endbibitem

\bibitem{Back}
\begin{botherref}
\oauthor{\bsnm{B{\"{a}}ck}, \binits{T.}},
\oauthor{\bsnm{B{\"{a}}ck}, \binits{T.}},
\oauthor{\bsnm{Rudolph}, \binits{G.}},
\oauthor{\bsnm{Schwefel}, \binits{H.-p.}}:
{Evolutionary Programming and Evolution Strategies: Similarities and
  Differences}.
IN PROCEEDINGS OF THE SECOND ANNUAL CONFERENCE ON EVOLUTIONARY PROGRAMMING,
11--22
\end{botherref}
\endbibitem

\bibitem{fortunato2010community}
\begin{barticle}
\bauthor{\bsnm{Fortunato}, \binits{S.}}:
\batitle{Community detection in graphs}.
\bjtitle{Physics reports}
\bvolume{486}(\bissue{3-5}),
\bfpage{75}--\blpage{174}
(\byear{2010})
\end{barticle}
\endbibitem

\bibitem{brandes2007finding}
\begin{bchapter}
\bauthor{\bsnm{Brandes}, \binits{U.}},
\bauthor{\bsnm{Delling}, \binits{D.}},
\bauthor{\bsnm{Gaertler}, \binits{M.}},
\bauthor{\bsnm{G{\"o}rke}, \binits{R.}},
\bauthor{\bsnm{Hoefer}, \binits{M.}},
\bauthor{\bsnm{Nikoloski}, \binits{Z.}},
\bauthor{\bsnm{Wagner}, \binits{D.}}:
\bctitle{On finding graph clusterings with maximum modularity}.
In: \bbtitle{International Workshop on Graph-Theoretic Concepts in Computer
  Science},
pp. \bfpage{121}--\blpage{132}
(\byear{2007}).
\bcomment{Springer}
\end{bchapter}
\endbibitem

\bibitem{newman2004fast}
\begin{barticle}
\bauthor{\bsnm{Newman}, \binits{M.E.}}:
\batitle{Fast algorithm for detecting community structure in networks}.
\bjtitle{Physical review E}
\bvolume{69}(\bissue{6}),
\bfpage{066133}
(\byear{2004})
\end{barticle}
\endbibitem

\bibitem{duch2005community}
\begin{barticle}
\bauthor{\bsnm{Duch}, \binits{J.}},
\bauthor{\bsnm{Arenas}, \binits{A.}}:
\batitle{Community detection in complex networks using extremal optimization}.
\bjtitle{Physical review E}
\bvolume{72}(\bissue{2}),
\bfpage{027104}
(\byear{2005})
\end{barticle}
\endbibitem

\bibitem{sherrington1975solvable}
\begin{barticle}
\bauthor{\bsnm{Sherrington}, \binits{D.}},
\bauthor{\bsnm{Kirkpatrick}, \binits{S.}}:
\batitle{Solvable model of a spin-glass}.
\bjtitle{Physical review letters}
\bvolume{35}(\bissue{26}),
\bfpage{1792}
(\byear{1975})
\end{barticle}
\endbibitem

\bibitem{boettcher2005extremal}
\begin{barticle}
\bauthor{\bsnm{Boettcher}, \binits{S.}}:
\batitle{Extremal optimization for sherrington-kirkpatrick spin glasses}.
\bjtitle{The European Physical Journal B-Condensed Matter and Complex Systems}
\bvolume{46}(\bissue{4}),
\bfpage{501}--\blpage{505}
(\byear{2005})
\end{barticle}
\endbibitem

\bibitem{khalil2017learning}
\begin{bchapter}
\bauthor{\bsnm{Khalil}, \binits{E.}},
\bauthor{\bsnm{Dai}, \binits{H.}},
\bauthor{\bsnm{Zhang}, \binits{Y.}},
\bauthor{\bsnm{Dilkina}, \binits{B.}},
\bauthor{\bsnm{Song}, \binits{L.}}:
\bctitle{Learning combinatorial optimization algorithms over graphs}.
In: \bbtitle{Advances in Neural Information Processing Systems},
pp. \bfpage{6348}--\blpage{6358}
(\byear{2017})
\end{bchapter}
\endbibitem

\bibitem{li2018combinatorial}
\begin{bchapter}
\bauthor{\bsnm{Li}, \binits{Z.}},
\bauthor{\bsnm{Chen}, \binits{Q.}},
\bauthor{\bsnm{Koltun}, \binits{V.}}:
\bctitle{Combinatorial optimization with graph convolutional networks and
  guided tree search}.
In: \bbtitle{Advances in Neural Information Processing Systems},
pp. \bfpage{539}--\blpage{548}
(\byear{2018})
\end{bchapter}
\endbibitem

\bibitem{kipf2016semi}
\begin{botherref}
\oauthor{\bsnm{Kipf}, \binits{T.N.}},
\oauthor{\bsnm{Welling}, \binits{M.}}:
Semi-supervised classification with graph convolutional networks.
arXiv preprint arXiv:1609.02907
(2016)
\end{botherref}
\endbibitem

\bibitem{halldorsson1997greed}
\begin{barticle}
\bauthor{\bsnm{Halld{\'o}rsson}, \binits{M.M.}},
\bauthor{\bsnm{Radhakrishnan}, \binits{J.}}:
\batitle{Greed is good: Approximating independent sets in sparse and
  bounded-degree graphs}.
\bjtitle{Algorithmica}
\bvolume{18}(\bissue{1}),
\bfpage{145}--\blpage{163}
(\byear{1997})
\end{barticle}
\endbibitem

\end{thebibliography}

\newcommand{\BMCxmlcomment}[1]{}

\BMCxmlcomment{

<refgrp>

<bibl id="B1">
  <title><p>Reducibility among combinatorial problems</p></title>
  <aug>
    <au><snm>Karp</snm><fnm>RM</fnm></au>
  </aug>
  <source>Complexity of computer computations</source>
  <publisher>Springer</publisher>
  <pubdate>1972</pubdate>
  <fpage>85</fpage>
  <lpage>-103</lpage>
</bibl>

<bibl id="B2">
  <title><p>Spin glass theory and beyond: An Introduction to the Replica Method
  and Its Applications</p></title>
  <aug>
    <au><snm>M{\'e}zard</snm><fnm>M</fnm></au>
    <au><snm>Parisi</snm><fnm>G</fnm></au>
    <au><snm>Virasoro</snm><fnm>M</fnm></au>
  </aug>
  <publisher>World Scientific Publishing Company</publisher>
  <pubdate>1987</pubdate>
  <volume>9</volume>
</bibl>

<bibl id="B3">
  <title><p>Modularity and community structure in networks</p></title>
  <aug>
    <au><snm>Newman</snm><fnm>ME</fnm></au>
  </aug>
  <source>Proceedings of the national academy of sciences</source>
  <publisher>National Acad Sciences</publisher>
  <pubdate>2006</pubdate>
  <volume>103</volume>
  <issue>23</issue>
  <fpage>8577</fpage>
  <lpage>-8582</lpage>
</bibl>

<bibl id="B4">
  <title><p>{Coordinate descent algorithms}</p></title>
  <aug>
    <au><snm>Wright</snm><fnm>SJ</fnm></au>
  </aug>
  <source>Mathematical Programming</source>
  <publisher>Springer Verlag</publisher>
  <pubdate>2015</pubdate>
  <volume>151</volume>
  <issue>1</issue>
  <fpage>3</fpage>
  <lpage>-34</lpage>
</bibl>

<bibl id="B5">
  <title><p>Particle swarm optimization</p></title>
  <aug>
    <au><snm>Kennedy</snm><fnm>J</fnm></au>
    <au><snm>Eberhart</snm><fnm>RC</fnm></au>
  </aug>
  <source>Proceedings of the IEEE International Conference on Neural
  Networks</source>
  <pubdate>1995</pubdate>
  <fpage>1942</fpage>
  <lpage>-1948</lpage>
</bibl>

<bibl id="B6">
  <title><p>Optimization by simulated annealing</p></title>
  <aug>
    <au><snm>Kirkpatrick</snm><fnm>S</fnm></au>
    <au><snm>Gelatt</snm><fnm>CD</fnm></au>
    <au><snm>Vecchi</snm><fnm>MP</fnm></au>
  </aug>
  <source>science</source>
  <publisher>American Association for the Advancement of Science</publisher>
  <pubdate>1983</pubdate>
  <volume>220</volume>
  <issue>4598</issue>
  <fpage>671</fpage>
  <lpage>-680</lpage>
</bibl>

<bibl id="B7">
  <title><p>Handbook of genetic algorithms</p></title>
  <aug>
    <au><snm>Davis</snm><fnm>L</fnm></au>
  </aug>
  <publisher>CUMINCAD</publisher>
  <pubdate>1991</pubdate>
</bibl>

<bibl id="B8">
  <title><p>Nature's way of optimizing</p></title>
  <aug>
    <au><snm>Boettcher</snm><fnm>S</fnm></au>
    <au><snm>Percus</snm><fnm>A</fnm></au>
  </aug>
  <source>Artificial Intelligence</source>
  <publisher>Elsevier</publisher>
  <pubdate>2000</pubdate>
  <volume>119</volume>
  <issue>1-2</issue>
  <fpage>275</fpage>
  <lpage>-286</lpage>
</bibl>

<bibl id="B9">
  <title><p>Fast local search for the maximum independent set
  problem</p></title>
  <aug>
    <au><snm>Andrade</snm><fnm>DV</fnm></au>
    <au><snm>Resende</snm><fnm>MG</fnm></au>
    <au><snm>Werneck</snm><fnm>RF</fnm></au>
  </aug>
  <source>Journal of Heuristics</source>
  <publisher>Springer</publisher>
  <pubdate>2012</pubdate>
  <volume>18</volume>
  <issue>4</issue>
  <fpage>525</fpage>
  <lpage>-547</lpage>
</bibl>

<bibl id="B10">
  <title><p>Automatic differentiation in PyTorch</p></title>
  <aug>
    <au><snm>Paszke</snm><fnm>A</fnm></au>
    <au><snm>Gross</snm><fnm>S</fnm></au>
    <au><snm>Chintala</snm><fnm>S</fnm></au>
    <au><snm>Chanan</snm><fnm>G</fnm></au>
    <au><snm>Yang</snm><fnm>E</fnm></au>
    <au><snm>DeVito</snm><fnm>Z</fnm></au>
    <au><snm>Lin</snm><fnm>Z</fnm></au>
    <au><snm>Desmaison</snm><fnm>A</fnm></au>
    <au><snm>Antiga</snm><fnm>L</fnm></au>
    <au><snm>Lerer</snm><fnm>A</fnm></au>
  </aug>
  <source>NIPS-W</source>
  <pubdate>2017</pubdate>
</bibl>

<bibl id="B11">
  <title><p>Simple statistical gradient-following algorithms for connectionist
  reinforcement learning</p></title>
  <aug>
    <au><snm>Williams</snm><fnm>RJ</fnm></au>
  </aug>
  <source>Machine learning</source>
  <publisher>Springer</publisher>
  <pubdate>1992</pubdate>
  <volume>8</volume>
  <issue>3-4</issue>
  <fpage>229</fpage>
  <lpage>-256</lpage>
</bibl>

<bibl id="B12">
  <title><p>Categorical Reparameterization with Gumbel-Softmax</p></title>
  <aug>
    <au><snm>Jang</snm><fnm>E</fnm></au>
    <au><snm>Gu</snm><fnm>S</fnm></au>
    <au><snm>Poole</snm><fnm>B</fnm></au>
  </aug>
  <source>5th International Conference on Learning Representations, {ICLR}
  2017, Toulon, France, April 24-26, 2017, Conference Track
  Proceedings</source>
  <publisher>OpenReview.net</publisher>
  <pubdate>2017</pubdate>
  <url>https://openreview.net/forum?id=rkE3y85ee</url>
</bibl>

<bibl id="B13">
  <title><p>The Concrete Distribution: {A} Continuous Relaxation of Discrete
  Random Variables</p></title>
  <aug>
    <au><snm>Maddison</snm><fnm>CJ</fnm></au>
    <au><snm>Mnih</snm><fnm>A</fnm></au>
    <au><snm>Teh</snm><fnm>YW</fnm></au>
  </aug>
  <source>5th International Conference on Learning Representations, {ICLR}
  2017, Toulon, France, April 24-26, 2017, Conference Track
  Proceedings</source>
  <pubdate>2017</pubdate>
  <url>https://openreview.net/forum?id=S1jE5L5gl</url>
</bibl>

<bibl id="B14">
  <title><p>{An Introduction to Continuous Optimization: Foundations and
  Fundamental Algorithms}</p></title>
  <aug>
    <au><snm>Andreasson</snm><fnm>N</fnm></au>
    <au><snm>Evgrafov</snm><fnm>A</fnm></au>
    <au><snm>Patriksson</snm><fnm>M</fnm></au>
  </aug>
  <pubdate>2007</pubdate>
  <fpage>400</fpage>
</bibl>

<bibl id="B15">
  <title><p>{Optimization of Complex Systems: Theory, Models, Algorithms and
  Applications}</p></title>
  <aug>
    <au><snm>Avraamidou</snm><fnm>S</fnm></au>
    <au><snm>Pistikopoulos</snm><fnm>EN</fnm></au>
  </aug>
  <source>Advances in Intelligent Systems and Computing</source>
  <publisher>Springer International Publishing</publisher>
  <pubdate>2020</pubdate>
  <volume>991</volume>
  <fpage>579</fpage>
  <lpage>-588</lpage>
  <url>http://link.springer.com/10.1007/978-3-030-21803-4</url>
</bibl>

<bibl id="B16">
  <title><p>{A Hybrid Simplex Search for Global Optimization with
  Representation Formula and Genetic Algorithm}</p></title>
  <aug>
    <au><snm>Zidani</snm><fnm>H</fnm></au>
    <au><snm>Ellaia</snm><fnm>R</fnm></au>
    <au><snm>Cursi</snm><fnm>ES</fnm></au>
  </aug>
  <source>Advances in Intelligent Systems and Computing</source>
  <publisher>Springer Verlag</publisher>
  <pubdate>2020</pubdate>
  <volume>991</volume>
  <fpage>3</fpage>
  <lpage>-15</lpage>
</bibl>

<bibl id="B17">
  <title><p>A Population-Based Stochastic Coordinate Descent Method</p></title>
  <aug>
    <au><snm>Rocha</snm><fnm>AMA</fnm></au>
    <au><snm>Costa</snm><fnm>MFP</fnm></au>
    <au><snm>Fernandes</snm><fnm>EM</fnm></au>
  </aug>
  <source>World Congress on Global Optimization</source>
  <pubdate>2019</pubdate>
  <fpage>16</fpage>
  <lpage>-25</lpage>
</bibl>

<bibl id="B18">
  <title><p>{A comparative study of population-based optimization algorithms
  for turning operations}</p></title>
  <aug>
    <au><snm>Yildiz</snm><fnm>AR</fnm></au>
  </aug>
  <source>Information Sciences</source>
  <publisher>Elsevier</publisher>
  <pubdate>2012</pubdate>
  <volume>210</volume>
  <fpage>81</fpage>
  <lpage>-88</lpage>
</bibl>

<bibl id="B19">
  <title><p>Gumbel-Softmax Optimization: A Simple General Framework for
  Combinatorial Optimization Problems on Graphs</p></title>
  <aug>
    <au><snm>Liu</snm><fnm>J</fnm></au>
    <au><snm>Gao</snm><fnm>F</fnm></au>
    <au><snm>Zhang</snm><fnm>J</fnm></au>
  </aug>
  <source>International Conference on Complex Networks and Their
  Applications</source>
  <pubdate>2019</pubdate>
  <fpage>879</fpage>
  <lpage>-890</lpage>
</bibl>

<bibl id="B20">
  <title><p>Graphical models, exponential families, and variational
  inference</p></title>
  <aug>
    <au><snm>Wainwright</snm><fnm>MJ</fnm></au>
    <au><snm>Jordan</snm><fnm>MI</fnm></au>
    <au><cnm>others</cnm></au>
  </aug>
  <source>Foundations and Trends{\textregistered} in Machine Learning</source>
  <publisher>Now Publishers, Inc.</publisher>
  <pubdate>2008</pubdate>
  <volume>1</volume>
  <issue>1--2</issue>
  <fpage>1</fpage>
  <lpage>-305</lpage>
</bibl>

<bibl id="B21">
  <title><p>{Evolutionary Programming and Evolution Strategies: Similarities
  and Differences}</p></title>
  <aug>
    <au><snm>B{\"{a}}ck</snm><fnm>T</fnm></au>
    <au><snm>B{\"{a}}ck</snm><fnm>T</fnm></au>
    <au><snm>Rudolph</snm><fnm>G</fnm></au>
    <au><snm>Schwefel</snm><fnm>Hp</fnm></au>
  </aug>
  <source>IN PROCEEDINGS OF THE SECOND ANNUAL CONFERENCE ON EVOLUTIONARY
  PROGRAMMING</source>
  <fpage>11</fpage>
  <lpage>---22</lpage>
  <url>http://citeseerx.ist.psu.edu/viewdoc/summary?doi=10.1.1.42.3637</url>
</bibl>

<bibl id="B22">
  <title><p>Community detection in graphs</p></title>
  <aug>
    <au><snm>Fortunato</snm><fnm>S</fnm></au>
  </aug>
  <source>Physics reports</source>
  <publisher>Elsevier</publisher>
  <pubdate>2010</pubdate>
  <volume>486</volume>
  <issue>3-5</issue>
  <fpage>75</fpage>
  <lpage>-174</lpage>
</bibl>

<bibl id="B23">
  <title><p>On finding graph clusterings with maximum modularity</p></title>
  <aug>
    <au><snm>Brandes</snm><fnm>U</fnm></au>
    <au><snm>Delling</snm><fnm>D</fnm></au>
    <au><snm>Gaertler</snm><fnm>M</fnm></au>
    <au><snm>G{\"o}rke</snm><fnm>R</fnm></au>
    <au><snm>Hoefer</snm><fnm>M</fnm></au>
    <au><snm>Nikoloski</snm><fnm>Z</fnm></au>
    <au><snm>Wagner</snm><fnm>D</fnm></au>
  </aug>
  <source>International Workshop on Graph-Theoretic Concepts in Computer
  Science</source>
  <pubdate>2007</pubdate>
  <fpage>121</fpage>
  <lpage>-132</lpage>
</bibl>

<bibl id="B24">
  <title><p>Fast algorithm for detecting community structure in
  networks</p></title>
  <aug>
    <au><snm>Newman</snm><fnm>ME</fnm></au>
  </aug>
  <source>Physical review E</source>
  <publisher>APS</publisher>
  <pubdate>2004</pubdate>
  <volume>69</volume>
  <issue>6</issue>
  <fpage>066133</fpage>
</bibl>

<bibl id="B25">
  <title><p>Community detection in complex networks using extremal
  optimization</p></title>
  <aug>
    <au><snm>Duch</snm><fnm>J</fnm></au>
    <au><snm>Arenas</snm><fnm>A</fnm></au>
  </aug>
  <source>Physical review E</source>
  <publisher>APS</publisher>
  <pubdate>2005</pubdate>
  <volume>72</volume>
  <issue>2</issue>
  <fpage>027104</fpage>
</bibl>

<bibl id="B26">
  <title><p>Solvable model of a spin-glass</p></title>
  <aug>
    <au><snm>Sherrington</snm><fnm>D</fnm></au>
    <au><snm>Kirkpatrick</snm><fnm>S</fnm></au>
  </aug>
  <source>Physical review letters</source>
  <publisher>APS</publisher>
  <pubdate>1975</pubdate>
  <volume>35</volume>
  <issue>26</issue>
  <fpage>1792</fpage>
</bibl>

<bibl id="B27">
  <title><p>Extremal optimization for Sherrington-Kirkpatrick spin
  glasses</p></title>
  <aug>
    <au><snm>Boettcher</snm><fnm>S</fnm></au>
  </aug>
  <source>The European Physical Journal B-Condensed Matter and Complex
  Systems</source>
  <publisher>Springer</publisher>
  <pubdate>2005</pubdate>
  <volume>46</volume>
  <issue>4</issue>
  <fpage>501</fpage>
  <lpage>-505</lpage>
</bibl>

<bibl id="B28">
  <title><p>Learning combinatorial optimization algorithms over
  graphs</p></title>
  <aug>
    <au><snm>Khalil</snm><fnm>E</fnm></au>
    <au><snm>Dai</snm><fnm>H</fnm></au>
    <au><snm>Zhang</snm><fnm>Y</fnm></au>
    <au><snm>Dilkina</snm><fnm>B</fnm></au>
    <au><snm>Song</snm><fnm>L</fnm></au>
  </aug>
  <source>Advances in Neural Information Processing Systems</source>
  <pubdate>2017</pubdate>
  <fpage>6348</fpage>
  <lpage>-6358</lpage>
</bibl>

<bibl id="B29">
  <title><p>Combinatorial optimization with graph convolutional networks and
  guided tree search</p></title>
  <aug>
    <au><snm>Li</snm><fnm>Z</fnm></au>
    <au><snm>Chen</snm><fnm>Q</fnm></au>
    <au><snm>Koltun</snm><fnm>V</fnm></au>
  </aug>
  <source>Advances in Neural Information Processing Systems</source>
  <pubdate>2018</pubdate>
  <fpage>539</fpage>
  <lpage>-548</lpage>
</bibl>

<bibl id="B30">
  <title><p>Semi-supervised classification with graph convolutional
  networks</p></title>
  <aug>
    <au><snm>Kipf</snm><fnm>TN</fnm></au>
    <au><snm>Welling</snm><fnm>M</fnm></au>
  </aug>
  <source>arXiv preprint arXiv:1609.02907</source>
  <pubdate>2016</pubdate>
</bibl>

<bibl id="B31">
  <title><p>Greed is good: Approximating independent sets in sparse and
  bounded-degree graphs</p></title>
  <aug>
    <au><snm>Halld{\'o}rsson</snm><fnm>MM</fnm></au>
    <au><snm>Radhakrishnan</snm><fnm>J</fnm></au>
  </aug>
  <source>Algorithmica</source>
  <publisher>Springer</publisher>
  <pubdate>1997</pubdate>
  <volume>18</volume>
  <issue>1</issue>
  <fpage>145</fpage>
  <lpage>-163</lpage>
</bibl>

</refgrp>
} 




\end{backmatter}
\end{document}